\documentclass[letterpaper]{article} 
\usepackage{aaai24}  
\usepackage{times}  
\usepackage{helvet}  
\usepackage{courier}  
\usepackage[hyphens]{url}  
\usepackage{graphicx} 
\usepackage{natbib}  
\usepackage{caption} 
\usepackage{algorithm}
\usepackage{algorithmic}
\usepackage{booktabs}
\usepackage{colortbl}  
\usepackage{xcolor}
\usepackage{multirow} 
\usepackage{subfig}
\usepackage{pifont}
\usepackage{amsmath}
\usepackage{amsfonts}
\usepackage{amssymb}
\definecolor{aliceblue}{rgb}{0.94, 0.97, 1.0}
\definecolor{deeppink}{RGB}{255,20,147}

\usepackage{hyperref}
\hypersetup{
    colorlinks = true, 
    urlcolor = deeppink, 
    linkcolor = black, 
    citecolor = black 
}

\usepackage{newfloat}
\usepackage{listings}
\DeclareCaptionStyle{ruled}{labelfont=normalfont,labelsep=colon,strut=off} 
\floatstyle{ruled}
\newfloat{listing}{tb}{lst}{}
\floatname{listing}{Listing}
\title{DGL: Dynamic Global-Local Prompt Tuning for Text-Video Retrieval} 
\author{
    Xiangpeng Yang\textsuperscript{\rm 1},
    Linchao Zhu\textsuperscript{\rm 2},
    Xiaohan Wang\textsuperscript{\rm 2},
    Yi Yang\textsuperscript{\rm 2} \thanks{Corresponding author} 
}

\affiliations{
    \textsuperscript{\rm 1} ReLER, AAII, University of Technology Sydney \\
    \textsuperscript{\rm 2}  CCAI, Zhejiang University \\

    Xiangpeng.Yang@student.uts.edu.au, wxh1996111@gmail.com, \{zhulinchao,yangyics\}@zju.edu.cn
}

\usepackage{bibentry}

\begin{document}

\maketitle

\begin{abstract}
Text-video retrieval is a critical multi-modal task to find the most relevant video for a text query. Although pretrained models like CLIP have demonstrated impressive potential in this area, the rising cost of fully finetuning these models due to increasing model size continues to pose a problem. To address this challenge, prompt tuning has emerged as an alternative. However, existing works still face two problems when adapting pretrained image-text models to downstream video-text tasks: (1) The visual encoder could only encode frame-level features and failed to extract global-level general video information. (2) Equipping the visual and text encoder with separated prompts failed to mitigate the visual-text modality gap. To this end, we propose \textit{\textbf{DGL}}, a cross-modal \textbf{D}ynamic prompt tuning method with \textbf{G}lobal-\textbf{L}ocal video attention. In contrast to previous prompt tuning methods, we employ the shared latent space to generate local-level text and frame prompts that encourage inter-modal interaction. Furthermore, we propose modeling video in a global-local attention mechanism to capture global video information from the perspective of prompt tuning. Extensive experiments reveal that when only \textbf{0.67\%} parameters are tuned, our cross-modal prompt tuning strategy DGL outperforms or is comparable to fully finetuning methods on MSR-VTT, VATEX, LSMDC, and ActivityNet datasets. \text{Code will be available at \url{https://github.com/knightyxp/DGL}}
\end{abstract}

\begin{figure}[h!]
  \centering
  \includegraphics[width=\linewidth]{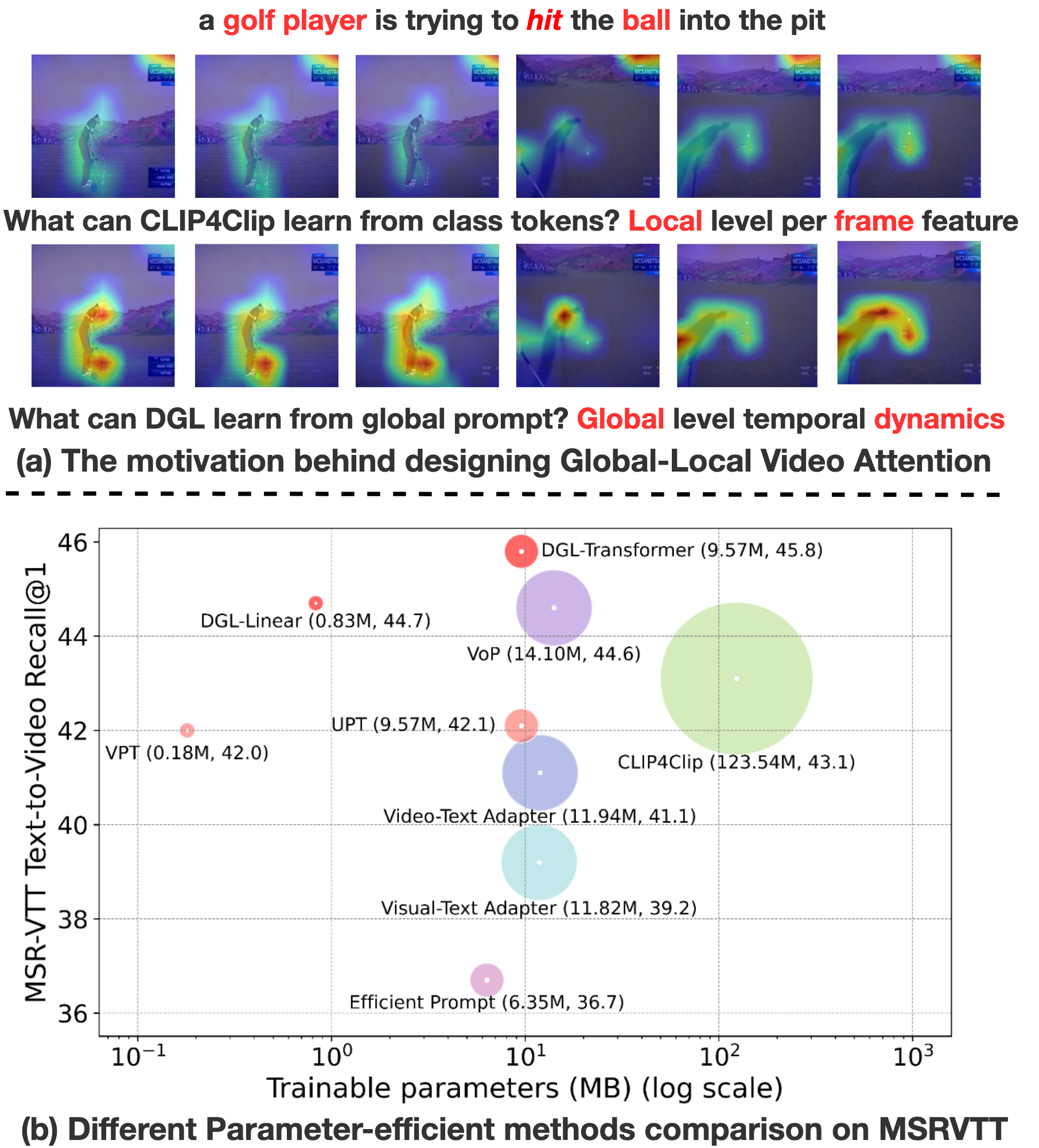}
  \caption {In Fig (a), we observed that frame attention methods, like CLIP4Clip, often emphasize non-semantic corners, missing the protagonist's action. This led us to design the global-local video attention for capturing global-level cross-frame dynamics. Fig (b) showcases a performance comparison on MSRVTT: DGL outperforms six PEFL methods and fine-tuned CLIP4Clip while updating minimal parameters.}
  \label{intro}
\end{figure}


\section{Introduction}

With the recent advancement of large-scale contrastive image-text pretraining methods \textit{i.e.,} CLIP \cite{radford2021learning}, the field of TVR (Text-Video Retrieval) has experienced many works \cite{luo2022clip4clip,gorti2022x,zhao2022centerclip,liu2022ts2,ma2022x,wang2022disentangled} to adapt image-text pretrained models like CLIP to the video-text domain and already achieve the promising performance. These approaches incur a large storage burden in actual scenarios because they need to store distinct new models for different tasks.
However, as the capacity of pretrained models is rapidly expanding nowadays, \textit{i.e.,} BEIT-3 \cite{wang2022image} has 1.9B parameters, and BLIP-L/14 has 578M parameters. Fully finetuning the entire model for each downstream task requires maintaining separate model weights for every dataset, hindering the feasibility of deployment given the growing model capacities.

To address this problem, inspired by the recent success of prompt tuning in both NLP \cite{lester2021power, liu2021p} and common visual recognition tasks \cite{jia2022visual}, we continue to introduce \textbf{prompt tuning} to the cross-modal domain. In this way, we only need to store the parameters of a few prompt vectors for various retrieval tasks and keep the pretrained model backbone frozen, thus reducing the total parameter cost.

Efficient Prompt \cite{ju2022prompting} is the first work that has attempted prompt tuning in this area, introducing learnable prompt vectors in the text input while viewing the video as separate frames. Despite incorporating an additional transformer for temporal encoding, the performance remains unsatisfactory. VoP \cite{Huang_2023_CVPR}, another recent prompt tuning approach, designs three kinds of video-specific prompts but optimizes the dual branches' prompts independently. We argue these current methods still fail to handle two key challenges when applying prompt tuning in text-video retrieval. (1) Cross-modal alignment: Existing schemes, like those in VoP and Efficient Prompt, optimize the two branches separately, making it challenging for the model to learn mutual cross-modal information effectively. (2) General video information extraction: Since CLIP is pretrained on image-text pairs, its primary focus is on local-level frame features rather than holistic, global-level video information. This inherent design leads to potential pitfalls when used directly for TVR tasks. In the top images of Fig \ref{intro} (a), the attention weights of CLIP4Clip's CLS tokens reveal this limitation. Specifically, the CLS tokens overlook the action of ``hit the ball" and instead allocate more attention to the upper-right corner -- a semantically void region.

To address the aforementioned issues, we propose dynamic global-local prompt tuning (coined as \textbf{DGL}) for text-video retrieval. Our approach generates dynamic local-level prompts (text and frame prompts) from a shared latent space. This allows for joint optimization and ensures the alignment of the two modalities. Moreover, we propose global-local video attention to model videos from both the global and local levels, capturing inter-frame temporal information with the global prompt and focusing on each frame's information with the local frame prompts. 


From a qualitative standpoint, the bottom image of Fig~{\ref{intro}}(a) clearly shows the effectiveness of \textbf{DGL}. In contrast to CLIP4Clip, our DGL can focus on the ``hit" action and the ball's trajectory into the pit. This demonstrates that our method can efficiently capture temporal dynamics. Besides, on the quantitative front, as shown in Fig~{\ref{intro}}(b), our method achieves the best trade-off between trainable parameters and performance. More specifically, with only 9.57M parameters updated, DGL achieves 45.8 R@1 on MSRVTT. These results demonstrate the importance of cross-modal interaction and a comprehensive understanding of video information. We undertake extensive experiments on four benchmarks, including MSR-VTT, VATEX, LSMDC, and ActivityNet. Our contributions can be summarized as follows:
\begin{itemize}
\item We propose to generate dynamic cross-modal prompts from the shared latent space to ensure the cross-modal interaction.

\item We propose a global-local attention mechanism for a comprehensive understanding of input video, facilitating effective learning of cross-frame temporal dynamics.

\item Compared to the fully finetuning CLIP4Clip and other prompt tuning methods, our DGL achieves superior or equivalent performance on R@1 across four public datasets while reducing 99.3\% of trainable parameters.
\end{itemize}

\section{Related Work}
\noindent\textbf{Text-Video Retrieval.} 
Text-video retrieval is a prevalent task in multimodal learning. Previous works like \cite{zhu2020actbert, wang2021t2vlad, sun2019videobert, bain2021frozen, lei2021less, liang2023local} utilize abundant video information for multimodal learning. With the pretrained models like CLIP \cite{radford2021learning} gaining traction, CLIP4Clip \cite{luo2022clip4clip} proposed to fine-tune CLIP on text-video retrieval by adding extensive similarity calculation mechanisms, which shows good performance on several benchmarks. This inspired follow-up research \cite{gorti2022x,bogolin2022cross,liu2022ts2,zhao2022centerclip,fang2021clip2video} which delved deeper into cross-modal learning. Recent research \cite{Wu_2023_CVPR, jin2023video, jin2023diffusionret} have introduced external tools for enhanced retrieval but predominantly utilize features extracted from CLIP. Our approach continues to build upon the foundation set by CLIP4Clip, emphasizing efficient parameter learning within the encoder.

\noindent\textbf{Parameter Efficient Methods.}
Fully fine-tuning is a common approach to adapting pretrained models into downstream tasks, but it can be inefficient due to large parameter sizes and time costs. To address this, parameter-efficient learning (PEFL) has been proposed, including adapter and prompt tuning methods. \textbf{Adapters} \cite{houlsby2019parameter} offers a plug-and-play approach by adding modules to pretrained networks. VL-Adapter \cite{sung2022vl} further extends the adapter to vision-and-language tasks.  Recently, \cite{jiang2022cross} introduced a weight-share mechanism and adopted the query-scoring frame features reweighting method proposed in \cite{bain2022clip} to boost performance. \cite{zhang2023multimodal} proposed a temporal adaptation and cross-modal interaction modules.
\textbf{Prompt tuning} \cite{lester2021power} is another parameter-efficient choice by introducing additional learnable parameters at the model's input. \cite{liu2021p} further applies prompts to each encoder layer for more knowledge probing. Adaptations to models like CLIP for specific tasks have been explored in \cite{zhou2022learning,zhou2022conditional}, with further refinements in image and cross-modal domains \cite{jia2022visual,zang2022unified,khattak2022maple}. In the text-video retrieval task, Efficient Prompt \cite{ju2022prompting} tried incorporating additional prompts into text queries but overlooked the potential of visual prompts in this context. While \cite {Huang_2023_CVPR} made advancements with video-specific prompts, they still hard to address cross-modal interactions in prompt tuning.  In this paper, we delve deeper into cross-modal prompt tuning and seek effective ways to represent videos, considering their inherent complexity compared to text.

\begin{figure*}[h]
  \centering
  \includegraphics[width=\linewidth]{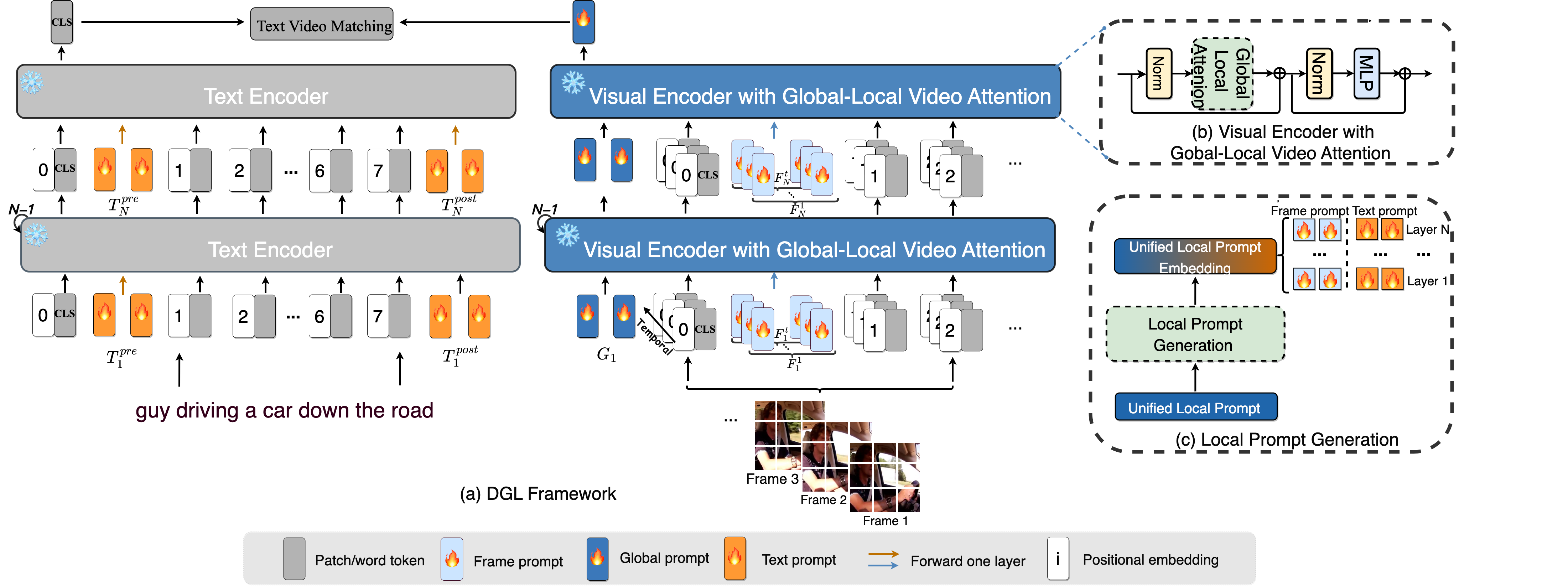}
  \caption{Overview of our Dynamic Global-Local prompt tuning Framework. DGL consists of the Local Prompt Generation, Text Encoder, and Visual Encoder. During downstream training, all the encoders are frozen, and only the parameter pictured with fire is trainable. The Local Prompt Generation ensures cross-modal interaction at the word-frame level, and the Global-Local Video Attention hints to the visual encoder to extract general video information from different perspectives.}
  \label {framework}
\end{figure*}

\section{Methods}
In this section, we will illustrate the details of our method. Firstly, we provide a comprehensive overview of the proposed DGL framework. Furthermore, we will introduce how to design global-local video attention to learn discriminative features from holistic video information.

\subsection{Global-Local Prompt Tuning Framework} \label{Framework}

Current parameter-efficient methods in TVR, such as Efficient Prompt \cite{ju2022prompting}, and VoP \cite{Huang_2023_CVPR} often neglect the crucial interaction between the visual and text modalities, as they focus only on inserting prompt vectors in the text input or prompting the dual branches separately. Additionally, deep prompts are necessary for more complex tasks like TVR, as evidenced by studies like VPT \cite{jia2022visual}, which showed shallow prompts to be less effective for traditional visual tasks like classification and segmentation. Meanwhile, classic fully finetuning methods like CLIP4Clip \cite{luo2022clip4clip}, TS2Net, Cap4Video, and HBI \cite{ liu2022ts2, Wu_2023_CVPR,jin2023video} still process videos as discrete frames, owing to CLIP's text-image pretraining. This complicates modeling inter-frame relationships and capturing temporal information.

To address these issues, we introduce DGL, a dynamic global-local prompt tuning method that facilitates global video-level information learning and ensures cross-modal alignment between frame-level visual features and word-level text features. Our DGL framework, as illustrated in Fig~{\ref{framework}}, consists of a local prompt generation module, a text encoder, and a visual encoder.

In the text branch, following \cite{ju2022prompting}, we design to learn a set of deep text prompts, including prefix text prompts ${T^{pre}_i} = \{T^{pre;j}_i \in \mathbb{R}^d | j \in \mathbb{N}, 1 \leq j \leq n_{pre}\}$ and postfix text prompts 
${T^{post}_i} = \{T^{post;j}_i \in \mathbb{R}^d | j \in \mathbb{N}, 1 \leq j \leq n_{post}\}$ for each layer index $i$. The prefix text prompts are added to the input text query before the word embedding, while the postfix text prompts are placed afterward. Here, $d$ is the dimension of the prompt vectors, $n_{pre}$ and $n_{post}$ denote the numbers of prefix and postfix prompts, respectively.

In the visual branch, to perform global-local video attention, we consider learning a single layer of $n_g$ global prompts ${G = \{G^j\in{\mathbb{R}}^{d} |j\in{\mathbb{N}}, 1\leq j\leq n_g \} }$ to capture the global information and a set of deep frame prompts $F_i^k = \{F_i^{k;j}\in{\mathbb{R}}^{d} |j\in{\mathbb{N}}, 1\leq k \leq t,1\leq j \leq n_f\}$ to extract frame information. Here, $k$ is frame index, $t$ is the number of frames, $j$ is the length index for frame prompts. $n_f$ and $n_g$ are the length of each frame prompts and global prompts.

 The visual encoder input contains global video prompts, frame patch tokens, and frame prompts. The text encoder input consists of text prompts and word tokens. Given the different choices of prompt generation modules, we dub our method as DGL-Transformer and DGL-Linear, respectively.

\subsection{Local Prompt Generation} \label{Local Prompt Generation}
We utilize two methods to generate local-level cross-modal prompts from the shared latent space and optimize them jointly. The first approach is the unified prompt transformer, and the second approach is the unified linear projection. We describe the details of these two methods as follows:

\textbf{Unified Prompt Transformer.} 
To exploit the cross-modal interaction at the fine-grained level, inspired by UPT \cite{zang2022unified}, we propose to generate frame prompts and text prompts from a  unified transformer (short as ``trans'').
For each layer, we merge text and visual prompts to form the unified prompt $U_i = [{T^{pre}_i},{T^{post}_i}, F_i^k]$, processed in a unified prompt transformer for cross-modal interaction. We learn layer-wise unified prompts ${U_i^{trans}}$ for both text and visual encoders. After transformation, ${U_i^{trans}}$ splits into three parts  ${\{T^{pre}_i,T^{post}_i, F_i^k\}}$, where the first two are sent into the text encoder and the last into the visual encoder. Notably, our unified prompt transformer only has a single layer. The hidden dimension matches the visual encoder's. Besides, we use an MLP Layer to adjust the text prompts' dimension.

\textbf{Unified Linear Projection.}
To further reduce the parameter cost, we consider using two simple linear layers, $U_{linear}^{pre}$ and $U_{linear}^{post}$, to map the frame prompts to the text prefix and text postfix prompts in each encoder layer respectively. This process can be formulated as follows:
\begin{equation}
   T^{pre}_i = U_{linear}^{pre}(F_i^k) \qquad 1\leq{i}\leq{N}
\end{equation}
\begin{equation}
   T^{post}_i = U_{linear}^{post}(F_i^k) \qquad 1\leq{i}\leq{N}
\end{equation}
Here $i$ is the layer index, $N$ is the total layers, and is 12 in our DGL, the same as the total encoder layers in CLIP. The length of each frame prompt and text prefix/postfix prompts are the same. All layers' multi-modal local prompts share the two projection layers. Therefore, the unified linear projection minimizes the parameter cost.

The two local prompt generation modules enable efficient interaction by mapping different modal prompts from the shared latent space (either the lightweight transformer or the linear layers), which ensures the cross-modal alignment between video frames and text words.

\subsection{Text Encoder and Visual Encoder} \label{text encoder and visual encoder}
\quad \textbf{Text Encoder.} \label{text encoder} In the text branch, combined with the text prefix prompts $T^{pre}_i$ and text postfix prompts $T^{post}_i$, we get the $i_{th}$ layer's text embedding as follows:
\begin{equation}
\begin{aligned}
\relax [ \, \underline{\hbox to 3mm{}}, W_i, \underline{\hbox to 3mm{}} \, ] = L_i^t([T^{pre}_{i-1}, W_{i-1}, T^{post}_{i-1}])
\end{aligned}
\end{equation}
where $[\cdot, \cdot, \cdot]$ refers to the concatenation operation, $L_i^t$ represents the $i_{th}$ text encoder layer, $W_i$ is the word embedding of $i_{th}$ text encoder layer. The prefix text prompts and postfix prompts are updated by the local prompt generation module in each layer. We get the final text representation by projecting from the final layer's word embedding $W_N$.

\textbf{Visual Encoder.} \label{visual encoder} For the $i_{th}$ ViT layer, combined with the global prompts ${G_i}$ and local frame prompts ${F_i^k}$, the prompt augmented ViT layer can be formulated as:
\begin{equation}
  [{{G_i},{C_i^k},{\,}{\underline{\hbox to 3mm{}}}{\,},{E_i^k}}] = L_i^v([{G_{i-1}},{C_{i-1}^k},F_{i-1}^{k} ,E_{i-1}^k])
\end{equation}
where $i$ is the layer index, and $k$ is the frame index. $L_i^v$ represents the $i_{th}$ visual encoder layer. $C_i^k$, $E_i^k$ represent each frame's [CLASS] embeddings and patch embeddings, respectively. Besides, since the global prompts ${G_i}$ are only prepended in the first visual layer, therefore, $\{G_i|i\neq 1 \} $ means the global prompt embedding for the $i_{th}$ ViT layer.

\begin{figure}[h]

 \includegraphics[width=\linewidth]{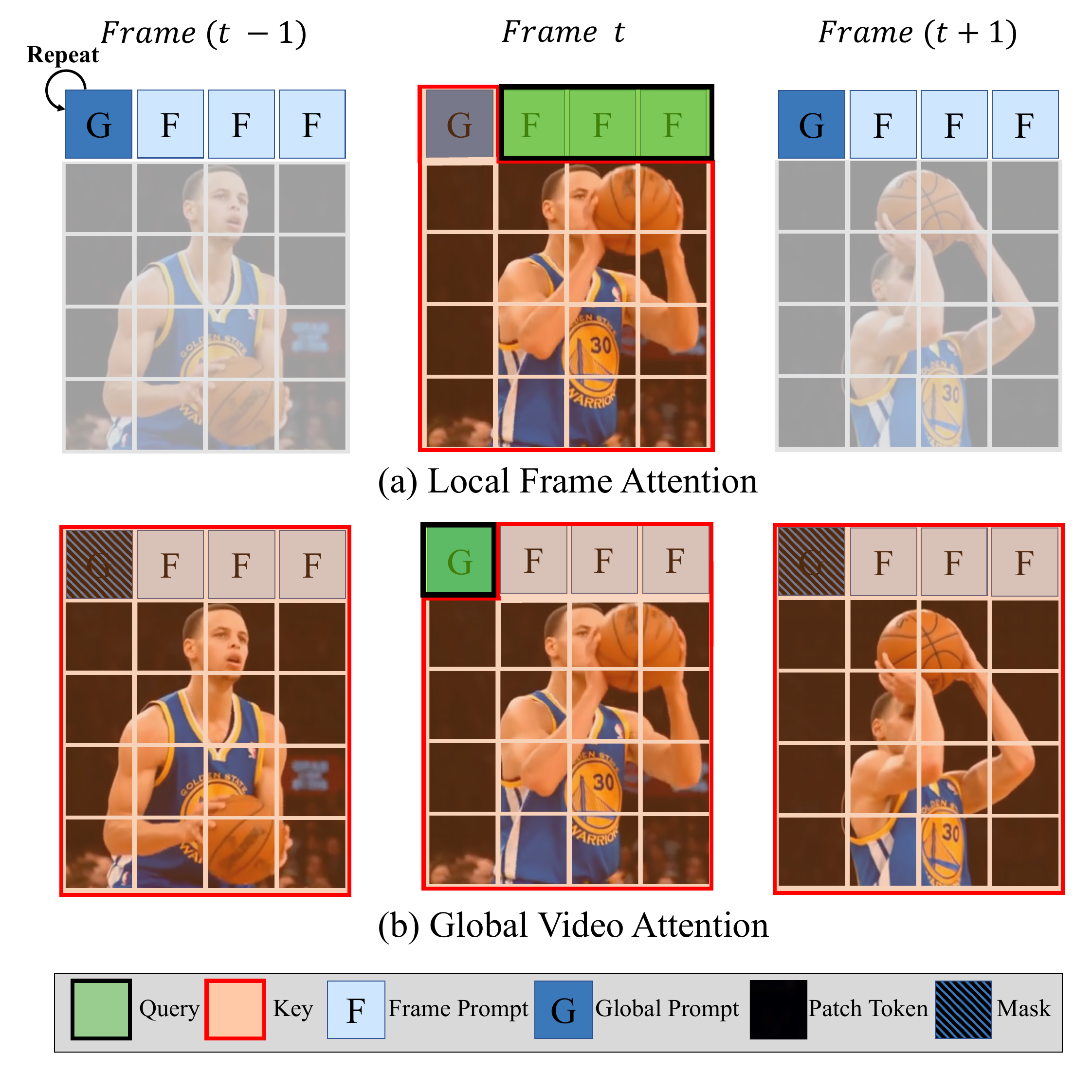}
\caption{Illustration of Global-Local Video Attention. The patches in the green mask serve as the queries in self-attention, and the patches in the orange mask are the key or value in self-attention. In the local frame attention, frame prompts serve as the query to investigate fine-grained local information in each frame; In global video attention, the global prompt acts as the query to excavate the global-level video information from all frames.}
\label {visualencoder}
\end{figure}

\subsection{Leveraging Global-Local Video Attention}  \label{Global-Local Attention}
In the text-video matching task, some text caption is related to single or short-latest frames. Thus the local frames feature is a must and basic. While some text query is a summary of behaviors of a video, therefore global video feature is also significant. Following \cite{xue2022clip}, we propose to devise local frame and global video attention in a share-parameter manner to extract frame and global video features based on the frame prompts and global prompts, respectively.

\textbf{Local Frame Attention.}
Specifically, for the local frame attention, we want each frame prompt could perceive each local frame information. As shown in Fig~{\ref{visualencoder}} (a), in the $i_{th}$ ViT layer, we concatenate the [CLASS] embeddings, frame prompt embeddings, and frame patch embeddings along the temporal dimension $k$, therefore, $ [C_{i-1}^k, F_{i-1}^k, E_{i-1}^k]$ serve as the query $Q^{loc}_{i-1}$. To ensure each frame prompt could perceive global information, we repeat global prompt embeddings $k$ times and concatenate them all. Therefore, we get $[G_{i-1}^k, C_{i-1}^k, F_{i-1}^{k}, E_{i-1}^k]$ as the key $K^{loc}_{i-1}$. Our Local frame attention can be formulated as follows:

\begin{equation}
 Q^{loc}_{i-1} = [C_{i-1}^k,F_{i-1}^k ,E_{i-1}^k]
\end{equation}
\begin{equation}
  K^{loc}_{i-1} = [G_{i-1}^k,C_{i-1}^k,F_{i-1}^{k},E_{i-1}^k]
\end{equation}
\begin{equation}
  [C_{i}^k,F_{i}^k ,E_{i}^k] = {Att(Q^{loc}_{i-1}, K^{loc}_{i-1},V^{loc}_{i-1})}
\end{equation}

\textbf{Global Video Attention.}
For global video attention, as shown in Fig~{\ref{visualencoder}} (b), global prompts $G_{i}$ need to learn global discriminant information. Therefore global prompts are attended to all frames' patch embeddings and prompt embeddings. This process can be formulated as follows:
\begin{equation}
 Q^{glo}_{i-1} = G_{i-1}
\end{equation}
\begin{equation} 
 {K^{glo}_{i-1}} = [G_{i-1},\cdots,X_{i-1}^k,V_{i-1}^k,E_{i-1}^k]
\end{equation}
\begin{equation}
  G_i = {Att}(Q^{glo}_{i-1},K^{glo}_{i-1},V^{glo}_{i-1})
\end{equation}
For each visual encoder layer, our local frame attention and global video attention are multi-head attention, using the CLIP pretrained visual encoder corresponding layer's parameter. But the query, key, and value are not the same. We perform the two attentions in query mode so that the query in the proposed mechanisms can perceive local frame-level and global video-level representations. Additionally, the two attention mechanisms are implemented in the sharing-parameter manner, which has two advantages: $\emph{(1)}$ Sharing parameter could reduce half parameters cost in the visual encoder. $\emph{(2)}$ Sharing parameter could excavate the pretrained CLIP visual encoder's potential extremely, which is validated in Table~{\ref{ablation}} (b).

\textbf{Similarity Calculation.}
Since the global prompts perceive each frame's information, we consider them a combination of fine-grained frames and global-discriminant video representations. Thus, we output the first global prompt, computing its similarity with the text representation.

Compared to the parameter-rich calculator, such as using four transformer layers to fuse temporal information in CLIP4Clip \cite{luo2022clip4clip}, our method is parameter-free in similarity computing. In addition, compared to the video feature re-weighting methods, like computing query-related frame features by cross-attention in \cite{gorti2022x}, through inner-product \cite{bain2022clip} or TopK \cite{liu2022ts2}, and then re-weighting the output video feature according to the frame-query similarity, our method is faster and query-agnostic. Especially in realistic applications, we could save much inference time because we only compute video and text features once. In contrast, the above re-weight methods need to be computed twice.

\textbf{Objective Function.}
In the training process, following \cite{luo2022clip4clip}, we still adopt symmetric cross-entropy loss. During downstream training, both the text and visual encoders are frozen. For various text-video retrieval scenarios, only the parameters of text/visual prompts and the local prompt generation module need to be stored. We only need to reuse a copy of the pretrained model (\textit{i.e.} CLIP), which reduces storage costs to the greatest extent.

\textbf{Discussion of other PEFL methods.} Adapters employ down-projection with nonlinear activation and up-projection mappings in each layer. However, adapters need a large intermediate compression dimension to maintain performance, undermining efficiency. Our test (Table \ref{msrvtt}) shows that adapter methods exceed the GPU memory of fully finetuned CLIP4Clip by over 50\%, using above 30GB against CLIP4Clip's 20.8GB. Also, these methods alter the original model's structure, complicating deployment. Therefore, we mainly focus on prompt tuning in this study.


\section{Experiments}

 \subsubsection{Datasets and Evaluation Metrics.} 
We conduct experiments on four datasets including MSR-VTT \cite{xu2016msr}, LSMDC \cite{rohrbach2015dataset}, ActivityNet \cite{heilbron2015activitynet} and VATEX \cite{wang2019vatex}. 

To measure the retrieval performance, we use standard metrics: recall at rank k (R@K, higher is better) and mean rank (MnR, lower is better). R@K computes the percentage of correct videos among the top K videos retrieved, we report the R@1, R@5, and R@10 results for each experiment. Mean rank computes the average rank of all correct answers. 
\subsubsection{Compared Baselines.}

We evaluate our approach against six strong baselines.

\textbf{Efficient Prompt} \cite{ju2022prompting} introduces prompt tuning in TVR by adding text prompts to text encoder input and a two-layer transformer for temporal modeling.

\textbf{VPT} \cite{jia2022visual} is a visual recognition method using prompt tuning, with VPT-deep showing notable results.

\textbf{UPT} \cite{zang2022unified} generates both visual and text prompts from a unified transformer layer. 

\textbf{Visual-Text Adapter}. Following \cite{houlsby2019parameter}, we add visual/text adapters after self-attention in each layer.

\textbf{Video-Text Adapter}. Based on Visual-Text Adapter, we replace the adapter in the visual encoder with ST-adapter \cite{pan2022st} to enhance the capability of extracting temporal information, which is inserted before multi-head attention.

\textbf{CLIP4CLip} \cite{luo2022clip4clip} is the fully finetuning baseline. We only compare the mean-pooling type for fairness, since our similarity calculator is also parameter-free.

\noindent \textbf{Implementation Details.} We use the CLIP (ViT-B/32) as the pre-trained model. During training, all the original parameters of CLIP are frozen unless explicitly mentioned. We apply a warm-up strategy followed by a cosine learning rate policy, using the AdamW optimizer with decoupled weight decay set to 0.2. The initial learning rate is 1e-2 for LSMDC and 5e-3 for the other three datasets. The max epochs are 10 for all datasets. Following CLIP4Clip, we uniformly sample 12 frames for MSRVTT, LSMDC, and VATEX and set the caption token length to 32. For ActivityNet, the frame length and caption length are set to 64. All the videos' short sides resize to 224, and the frame per second (fps) is 3. By default, the lengths of the frame prompts, text prefix/postfix prompts, and global prompts are all set to 4. Also, the depth of frame prompts and text prefix/postfix prompts is set to 12 by default. The inner dim of the adapter is set to 368. All experiments are done with mixed precision.

\subsection{Results on Benchmarks}

\definecolor{mygray}{gray}{0.95}

\begin{table*}[h!]
\renewcommand\arraystretch{1.1}
\resizebox{\linewidth}{!}
{
\begin{tabular}{llllllllllll}
\toprule[1.5pt]
\multicolumn{2}{c}{} & Trainable & Memory & \multicolumn{4}{c}{Text → Video} & \multicolumn{4}{c}{Video → Text} \\
Type & Methods &  Params(MB) ↓ & Usage(GB) ↓ & R@1↑ & R@5↑ & R@10↑ & MnR↓ & R@1↑ & R@5↑ & R@10↑ & MnR↓ \\ 
\midrule
\rowcolor{mygray}\multicolumn{12}{l}{\emph{CLIP-ViT-B/32}} \\
\multicolumn{1}{c}{\multirow{1}{*}{{Finetune}}} &
  \multicolumn{1}{l|}{CLIP4Clip} &
  123.54 &
  \multicolumn{1}{l|}{20.80} &
  43.1 &
  70.4 &
  80.8 &
  \multicolumn{1}{l|}{16.2} &
  43.1 &
  70.5 &
  81.2 &
  12.4 \\ 

\hline
{\multirow{2}{*}{Adapter}} &
  \multicolumn{1}{l|}{Visual-Text Adapter} &
  11.82 &
  \multicolumn{1}{l|}{30.71} &
  39.2 &
  65.7 &
  76.1 &
  \multicolumn{1}{l|}{17.6} &
  40.7 &
  68.8 &
  77.6 &
  13.7 \\
 &
  \multicolumn{1}{l|}{Video-Text Adapter} &
  11.94 &
  \multicolumn{1}{l|}{31.59} &
  41.1 &
  67.0 &
  77.1 &
  \multicolumn{1}{l|}{17.4} &
  42.6 &
  68.4 &
  78.4 &
  13.8 \\ \hline
\multicolumn{1}{c}{\multirow{7}{*}{Prompt}} &
  \multicolumn{1}{l|}{Efficient Prompt \cite{ju2022prompting}} &
  6.35 &
  \multicolumn{1}{l|}{-} &
  36.7 &
  64.6 &
  -   &
  \multicolumn{1}{l|}{-} &
  -    &
  -    &
  -    &
  - \\
\multicolumn{1}{c}{} &
  \multicolumn{1}{l|}{VPT \cite{jia2022visual}} &
  \textbf{0.18} &
  \multicolumn{1}{l|}{20.98} &
  42.0 &
  66.6 &
  77.3 &
  \multicolumn{1}{l|}{19.2} &
  39.4 &
  66.8 &
  77.2 &
  16.2 \\
\multicolumn{1}{c}{} &
  \multicolumn{1}{l|}{UPT \cite{zang2022unified}} &
  9.57 &
  \multicolumn{1}{l|}{23.46} &
  42.1 &
  67.7 &
  78.2 &
  \multicolumn{1}{l|}{16.5} &
  42.6 &
  70.3 &
  79.3 &
  12.3 \\
\multicolumn{1}{c}{} &
  \multicolumn{1}{l|}{{VoP$^{F+C}$} \cite{Huang_2023_CVPR}} &
  {14.10} &
  \multicolumn{1}{l|}{-} &
  44.6 &
  69.9 &
  80.3 &
  \multicolumn{1}{l|}{16.3} &
  44.5 &
  70.7 &
  80.6 &
  11.5 \\

\multicolumn{1}{c}{} &
  \multicolumn{1}{l|}{DGL-Linear(Ours)} &
  0.83 &
  \multicolumn{1}{l|}{\textbf{18.75}} &
  44.7 &
  70.5 &
  79.2 &
  \multicolumn{1}{l|}{16.2} &
  42.1 &
  70.0 &
  80.6 &
  13.4 \\  
  
 \multicolumn{1}{c}{} &
  \multicolumn{1}{l|}{DGL-Transformer(Ours)} &
  9.57 &
  \multicolumn{1}{l|}{20.69} &
  45.8 &
  69.3 &
  79.4 &
  \multicolumn{1}{l|}{16.3} &
  43.5 &
  70.5 &
  80.7 &
  13.1 \\ 
\multicolumn{1}{c}{} &
  \multicolumn{1}{l|}{+ QB-Norm\cite{bogolin2022cross}} &
  9.57 &
  \multicolumn{1}{l|}{20.69} &
  47.0 &
  70.4 &
  81.0 &
  \multicolumn{1}{l|}{16.4} &
  44.9 &
  70.7 &
  79.6 &
  13.3 \\ 
\hline
\rowcolor{mygray}\multicolumn{12}{l}{\emph{CLIP-ViT-B/16}} \\
\multicolumn{1}{c}{\multirow{4}{*}{}} &
\multicolumn{1}{l|} {CLIP4Clip + ViT-B/16}  &
    123.54 &
  \multicolumn{1}{l|}{25.70} &
  45.6 &
  71.2 &
  80.9 &
  \multicolumn{1}{l|}{15.2} &
  43.2 &
  72.5 &
  80.7 &
  10.9  \\
  & \multicolumn{1}{l|}{{VoP$^{F+C}$} + ViT-B/16 \cite{Huang_2023_CVPR}} &{14.10} &
  \multicolumn{1}{l|}{-} &
  47.7 &
  72.4 &
  82.2 &
  \multicolumn{1}{l|}{12.0} &
  - &
  - &
  - &
  - \\
& \multicolumn{1}{l|} {\textbf{DGL-Linear(Ours)} + ViT-B/16 }  &
   0.83 &
\multicolumn{1}{l|}{22.86} &
   48.3 &
   71.8 &
   80.6 &
\multicolumn{1}{l|}{13.4} & 
   45.7 &
   74.0 &
   82.9 &
   10.9 \\  
\rowcolor{gray!30}  &
\multicolumn{1}{l|} {+ QB-Norm\cite{bogolin2022cross}}  &
   0.83 &
\multicolumn{1}{l|}{22.86} &
   \textbf{49.7} &
   73.1 &
   82.3 &
\multicolumn{1}{l|}{15.1} & 
  \textbf{47.8}&
   74.1 &
   83.3 &
   10.6 \\   
\bottomrule[1.5pt]
\end{tabular}
}
\caption{Retrieval results on the MSR-VTT-9K dataset.}
\label{msrvtt}
\end{table*}

\begin{table*}[t]
\renewcommand\arraystretch{1.1}
\resizebox{\linewidth}{!}{
\begin{tabular}{llllllllllllll}
\toprule[1.5pt]
\multicolumn{2}{c}{} &
  \multicolumn{4}{c}{VATEX} &
  \multicolumn{4}{c}{LSMDC} &
  \multicolumn{4}{c}{ActivityNet} \\ 
\multicolumn{1}{c}{Type} &
  \multicolumn{1}{l}{Methods} &
  \multicolumn{1}{c}{R@1↑} &
  \multicolumn{1}{c}{R@5↑} &
  \multicolumn{1}{c}{R@10↑} &
  \multicolumn{1}{c}{MnR↓} &
  \multicolumn{1}{c}{R@1↑} &
  \multicolumn{1}{c}{R@5↑} &
  \multicolumn{1}{c}{R@10↑} &
  \multicolumn{1}{c}{MnR↓} &
  \multicolumn{1}{c}{R@1↑} &
  \multicolumn{1}{c}{R@5↑} &
  \multicolumn{1}{c}{R@10↑} &
  \multicolumn{1}{c}{MnR↓} 
  \\ 
\midrule
Finetune & \multicolumn{1}{l|} {CLIP4Clip} &  55.9 & 89.2 & {95.0} & \multicolumn{1}{l|} {{3.9}} &  20.7 & 38.9 & 47.2 &\multicolumn{1}{l|} {65.3}& {40.5} & {72.4} & - &{7.5} \\ 
\hline
\multicolumn{1}{c}{\multirow{2}{*}{Adapter}} & \multicolumn{1}{l|}{Visual-Text Adapter}  & 53.1 & 85.0 & 92.3 & \multicolumn{1}{l|} {4.9}  & 18.0 & 34.4 & 43.5 &\multicolumn{1}{l|} {75.2}& 33.5 & 64.8 & 77.5 &10.9 \\
& \multicolumn{1}{l|}{Video-Text Adapter}  & 53.5 & 85.0 & 92.4 & \multicolumn{1}{l|} {4.7} & 18.3 & 35.5 & 44.0 & \multicolumn{1}{l|}{74.8}& 36.4 & 66.1 & 79.6 &10.0 \\ 
\hline
\multicolumn{1}{c}{\multirow{5}{*}{Prompt}}& \multicolumn{1}{l|} {Efficient Prompt \cite{ju2022prompting}}  & - & - &  - & \multicolumn{1}{l|} {-} & 13.4 & 29.5 &  - & \multicolumn{1}{l|} {-} & - & - & - &- \\
& \multicolumn{1}{l|} {{VoP$^{F+P}$} \cite{Huang_2023_CVPR}}  & - & - &  - & \multicolumn{1}{l|} {-} & 20.7 & {40.7} & {49.7} & \multicolumn{1}{l|} {{59.1}} & 36.1 & 65.5 & 78.5 &10.9 \\

& \multicolumn{1}{l|} {DGL-Transformer(Ours)} & 54.3 & 85.5 & 92.3 & \multicolumn{1}{l|} {4.9} & 21.2 & 37.8 & {48.8} & \multicolumn{1}{l|} {66.5} & 40.1 & 69.5 & 80.9 &9.1\\
 & \multicolumn{1}{l|} {\textbf{DGL-Linear(Ours)}} & {56.2} &  87.1 &  {93.5} & \multicolumn{1}{l|} {4.1} & {21.4} & {39.4} & 48.4 &  \multicolumn{1}{l|} {{64.3}} & 38.6 & 69.2 & {81.6} &9.0 \\
\rowcolor{gray!30}  &  \multicolumn{1}{l|} {+ QB-Norm\cite{bogolin2022cross}} & \textbf{57.3} & 87.0 & 93.3 & \multicolumn{1}{l|} {4.2} & \textbf{21.6} & 39.3 & {49.0} &  \multicolumn{1}{l|}{64.4} & \textbf{43.1} & {72.3} & {82.7} & {8.6}
\\
\bottomrule[1.5pt]
\end{tabular}
}
\caption{Combined Retrieval Results for VATEX, LSMDC, and ActivityNet Datasets.}
\label{combined_results}
\end{table*}


\textbf{Results on MSR-VTT.} 
Table~{\ref{msrvtt}} presents MSRVTT-9K results. With only 0.83MB parameters trainable, DGL-Linear (ViT-B/16) enhances R@1 by 2.7\% over CLIP4Clip. For ViT-B/32, DGL-Transformer tops the list and exceeds Efficient Prompt and VoP in $T{\rightarrow}V$ R@1 by 9.1\% and 1.2\%. Across the board, DGL surpasses all adapters and prompt methods. Notably, DGL-Linear consumes just 18.75 GB of GPU memory, less than CLIP4Clip's 20.8 GB.  

\noindent \textbf{Results on the other three datasets.}
Table~{\ref{combined_results}} displays the retrieval results on VATEX, LSMDC, and ActivityNet. For ViT-B/32, tuning only 0.83MB parameters, DGL surpasses CLIP4Clip by 0.3\% and 0.7\% in $T{\rightarrow}V$ R@1 on VATEX and LSMDC and comparable performance on ActivityNet. This underscores our method's efficiency. Notably, we outperform Efficient Prompt by 8.0\% in LSMDC's $T{\rightarrow}V$ R@1 and achieve a 4.0\% lead over VoP on ActivityNet.

\begin{table*}[t]
\small
\begin{minipage}[b]{0.25 \textwidth}
\centering
\subfloat[\centering {Comparison of different visual output. ``GL" indicates Global-Local. }]
{
\setlength{\tabcolsep}{0.8pt}
\begin{tabular}{l|llll}
\toprule[1.25pt]
Visual Output & R@1$\uparrow$ & R@5$\uparrow$ & R@10$\uparrow$  & MnR$\downarrow$ \\   \hline
\multicolumn{1}{l|}{} & \multicolumn{4}{c}{\cellcolor{mygray} \emph{Text → Video}} \\  
\rowcolor{gray!30}  \small{First Global Prompt}         & \textbf{45.8} & 69.3          & 79.4    & 16.3       \\      %
\small{Avg Global Features }       & 43.5          & 69.7 & 79.7   & 16.9       \\   %
\smash{Avg Local Features}        & 41.5          & 68.3          & 77.2     & 16.0             \\  %
\small{Avg GL Features}  & 42.5  & 68.8   & 77.6  &15.6  \\ \hline %
\multicolumn{1}{l|}{} & \multicolumn{4}{c}{\cellcolor{mygray} \emph{Video → Text}} \\  
\small{First Global Prompt}        & 43.5          & 70.5     & 80.7  & 13.1    \\ %
\small{Avg Global Features}      & 43.1    & 70.0          & 79.9  & 12.5  \\ %
\small{Avg  Local Features}        & 41.9          & 70.9 & 79.3 & 12.4  \\     %
\small{Avg GL Features}    & 44.3 & 69.6     & 79.9  & 12.4  \\    %
\bottomrule[1.25pt]
\end{tabular}
}
\end{minipage}
\hspace{0.11\textwidth}
\begin{minipage}[b]{0.25 \textwidth}
\centering
\subfloat[Ablation of global-local video attention \label{new attention mechanisms}]{
\setlength{\tabcolsep}{0.8pt}
\begin{tabular}{lll|cccc}
\toprule[1.25pt]
 global & local  & share  & R@1↑ & R@5↑ & \small{R@10↑}  & MnR$\downarrow$ \\ \midrule
\multicolumn{3}{l|}{} & \multicolumn{4}{c}{\cellcolor{mygray} \emph{Text → Video}} \\  
\ding{51}     &    &         & 40.4    & 67.8    & 77.3   & 18.3     \\  %
& \ding{51}     &      & 42.0 & 68.7 & 78.2  & 16.9 \\  %

\ding{51}        & \ding{51}     &      & 43.5 & 70.1 & 79.0 &17.1   \\  %
\rowcolor{gray!30}  \ding{51}        & \ding{51}     & \ding{51}        & \textbf{45.8} & 69.3 & 79.4  & 16.3 \\ \hline  %
\multicolumn{3}{l|}{} & \multicolumn{4}{c}{\cellcolor{mygray} \emph{Video → Text}} \\  
\ding{51}   &         &        & 41.3    & 67.8    & 76.8 & 14.3  \\  %
& \ding{51}    &      & 42.2 & 69.7 & 78.6  & 12.6 \\   %
\ding{51}       & \ding{51}      &      & 43.1 & 69.4 & 80.0 & 13.6 \\   %

\ding{51}        & \ding{51}     & \ding{51} & 43.5 & 70.5 & 80.7 & 13.1  \\    %
\bottomrule[1.25pt]
\end{tabular}
}
\end{minipage}
\hfill
\begin{minipage}[b]{0.3 \textwidth}
\centering
\subfloat[{\centering Effect of text prompt position, ``UP" indicates trainable parameters. \label{text prompt position}}]{
\setlength{\tabcolsep}{0.8pt}
\begin{tabular}{l|cccc}
\toprule[1.25pt]
Position &   R@1$\uparrow$ & R@5$\uparrow$ & R@10$\uparrow$ & MnR$\downarrow$ \\
\midrule
\multicolumn{1}{l|}{} & \multicolumn{4}{c}{\cellcolor{mygray} \emph{Text → Video}} \\ 
4+X       & 43.9          & 70.0          & 79.0          & 16.1           \\ 
8+X      & 45.0      & 70.2 & 80.3 & 15.0 \\  
\rowcolor{gray!30} 4+X+4            & \textbf{45.8} & 69.3          & 79.4          & 16.3          \\  \hline
\multicolumn{1}{l|}{} & \multicolumn{4}{c}{\cellcolor{mygray} \emph{Video → Text}} \\  
4+X           & 42.4          & 70.5          & 80.3          & 12.8    \\ 
8+X       & 43.0          & 70.1          & 81.3 & 12.4 \\  
 4+X+4             & 43.5 & 70.5 & 80.7          & 13.1      \\ 
\bottomrule[1.25pt]
\end{tabular}
}
\end{minipage}
\begin{minipage}[b]{0.2\textwidth}
\centering
\subfloat[\centering {Verifying DGL on other baselines, ``*" indicates freeze backbone.}]
{
\setlength{\tabcolsep}{0.8pt}
\begin{tabular}{l>{\centering\arraybackslash}m{1cm}|ccc}
\toprule[1.25pt]
& &  \multicolumn{3}{c} {Text → Video } \\
Methods & \small{UP(M)}↓ & R@1↑ & R@5↑ & R@10↑      \\ \midrule
\rowcolor{mygray}\multicolumn{5}{l}{\emph{BLIP\cite{li2022blip}}} \\
\small{Full\cite{liu2022tokenmix}}  & 226.51  & 47.6   &  73.4 & 81.8    \\ 
\small{Token Mix}  & 7.07  & 47.1   & 70.8  & 80.5        \\  
\rowcolor{gray!30} DGL(ours)    & \textbf{0.30} &   \textbf{48.6} & 71.4 & 79.7 \\ \hline  
\rowcolor{mygray}\multicolumn{5}{l}{\emph{X-CLIP\cite{ma2022x}}} \\ 
X-CLIP* & 8.0   & 39.6   &  66.8 & 76.4    \\ 
\rowcolor{gray!30} \small {+DGL-Linear} & \textbf{2.9}  & \textbf{44.0}   & 69.9 & 79.6   \\
\bottomrule[1.25pt]
\end{tabular}
}
\end{minipage}
\hspace{0.16\textwidth}
\begin{minipage}[b]{0.25\textwidth}
\centering
\subfloat[{\centering Abalation experiment of DGL-Linear projection direction. \label{projection direction}}]{
\setlength{\tabcolsep}{0.8pt}
\begin{tabular}{l|cccp{0.7cm}}
\toprule[1.25pt]
Direction & R@1$\uparrow$ & R@5$\uparrow$ & R@10$\uparrow$ & \small{MnR$\downarrow$} \\
\midrule
\multicolumn{1}{l|}{} & \multicolumn{4}{c}{\cellcolor{mygray} \emph{Text → Video}} \\ 
T→V & 43.4  & 69.6  & 79.7 & 16.2  \\
\rowcolor{gray!30} V→T  & \textbf{44.7} & 70.5 & 79.2 & 16.2 \\
\hline
\multicolumn{1}{l|}{} & \multicolumn{4}{c}{\cellcolor{mygray} \emph{Video → Text}} \\  
T→V &  43.0 & 70.0  & 79.5  & 12.6 \\
V→T &  42.1  &70.0 &80.6 & 13.4 \\
\bottomrule[1.25pt]
\end{tabular}
}
\end{minipage}
\hspace{0.02\textwidth}
\begin{minipage}[b]{0.3\textwidth}
\centering
\subfloat[{\centering Effect of generating cross-modal prompts from the shared latent space. \label{prompt generation}}]{
\setlength{\tabcolsep}{0.8pt}
\begin{tabular}{l|cccp{0.7cm}}
\toprule[1.25pt]
Method  &    R@1$\uparrow$ & R@5$\uparrow$ & R@10$\uparrow$ & \small{MnR$\downarrow$}  \\ \midrule
\multicolumn{1}{l|}{} & \multicolumn{4}{c}{\cellcolor{mygray} \emph{Text → Video}} \\ 
Baseline    & 43.8   & 68.7  & 80.2 & 16.2 \\
\rowcolor{gray!30}  DGL-Transformer      & \textbf{45.8}  & 69.3    & 79.4    & 16.3    \\ \hline
\multicolumn{1}{l|}{} & \multicolumn{4}{c}{\cellcolor{mygray} \emph{Video → Text}} \\  
Baseline    & 43.9  & 69.4     & 80.1    & 12.2 \\
DGL-Transformer     & 43.5    & 70.5    & 80.7    & 13.1    \\ \bottomrule[1.25pt]
\end{tabular}
}
\end{minipage}

\caption{Ablation studies on the MSRVTT-9K dataset}
\label{ablation}
\end{table*}

\subsection{Ablation Study}
In this section, we thoroughly ablate DGL on MSR-VTT-9K using DGL-Transformer (ViT-B/32) unless specified.

\begin{figure}[H]
  \centering
  \includegraphics[width=\linewidth]{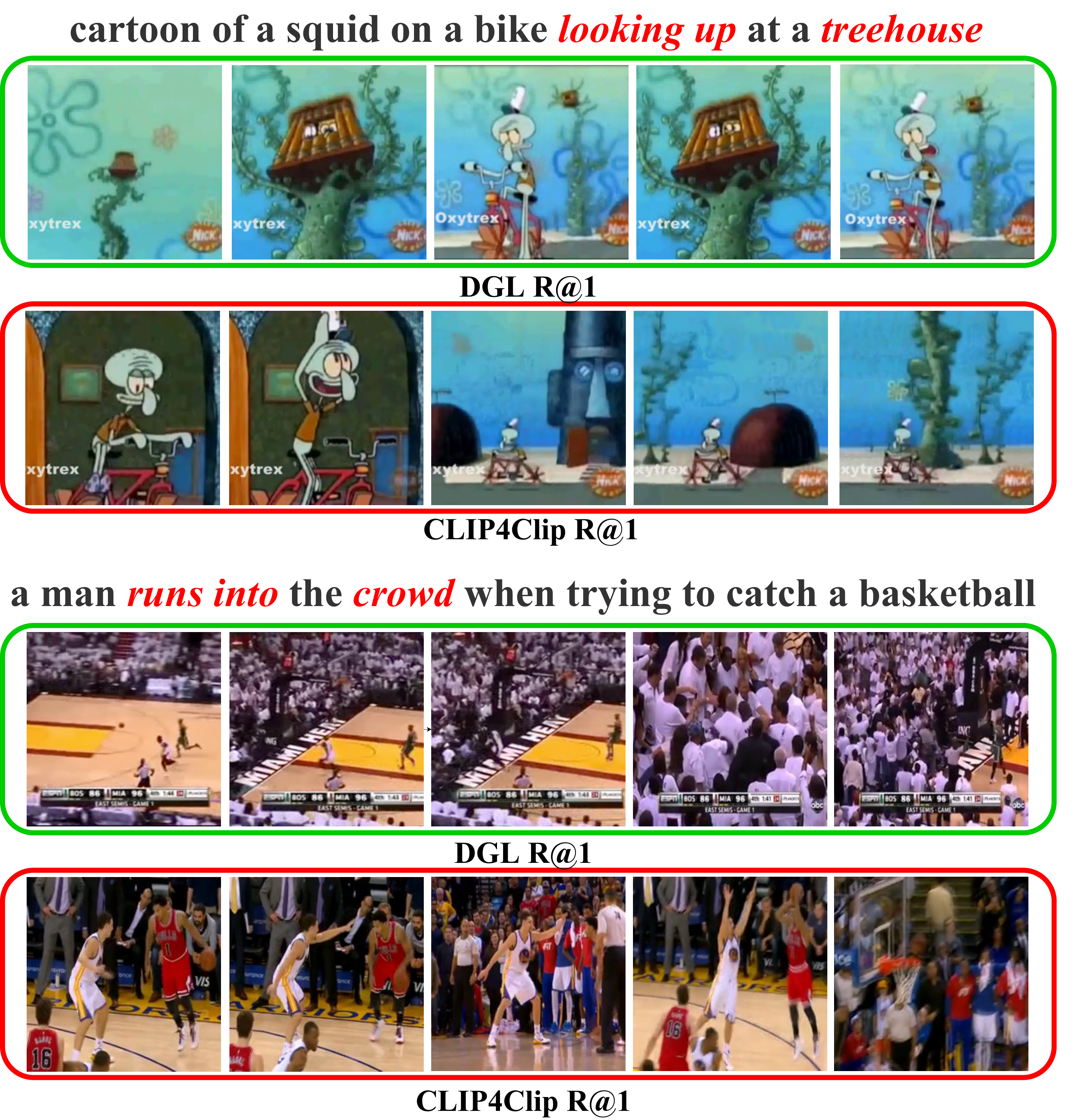}
  \caption{Visualization of text-video retrieval results. Frames in the green box are DGL R@1 results, while those in the red box are CLIP4clip R@1 results.}
  \label{visualization}
\end{figure}

\textbf{The output of the visual encoder.} We compare four different types of visual encoder output as the final video representation as shown in Table~{\ref{ablation}} (a). We found that using the first global prompt performs best, this is because our attention is designed for the global prompt, which can perceive global information and local frame details.

\textbf{Ablation of global-local video attention.} We assess global-local video attention by testing global/local attention and shared parameters separately. Table~{\ref{ablation}} (b) shows that global-local video attention with sharing parameters outperforms the other methods, demonstrating that both global and local information is crucial for text-video retrieval.

\textbf{Effect of the postfix text prompt.} Following Efficient Prompt \cite{ju2022prompting}, which adds text prefix and postfix prompts only in the input layer, we extend this to all encoder layers. Table~{\ref{ablation}} (c) shows [4+X+4] deep text prompts outperforming [8+X] or [4+X], maximizing prompt potential.

\textbf{Verify DGL on other baselines.} We evaluated DGL on different structures and other CLIP-based methods. \textbf{(1)} Following Token Mix \cite{liu2022tokenmix}, we integrated DGL with BLIP (ViT-B/16) \cite{li2022blip}, applying global-local video attention in the frozen visual encoder. Table~{\ref{ablation}} (d) top part shows that DGL surpasses the fully finetuning/PEFL method. \textbf{(2)} For CLIP-based method comparison, we focus on parameter-efficient designs and compare with X-CLIP \cite{ma2022x} by freezing the CLIP backbone for fairness. The bottom part demonstrates DGL's effectiveness.

\textbf{Why project Linear from visual to text?} Visual features are more complex than textual, as videos typically contain more information. Projecting simpler text to complex visual features is challenging. Table~{\ref{ablation}} (e) shows Visual2Text projection achieves higher R@1, validating our claim.


\textbf{Generating from the shared latent space.}
DGL-Transformer enhances cross-modal interaction and local consistency by generating prompts from a shared latent space. Compared with the divided prompt baselines while maintaining global-local video attention, the 2\% T→V R@1 improvement in Table~{\ref{ablation}} (f) demonstrates its effectiveness.


\textbf{Retrieval results comparison.}
In Fig~{\ref{visualization}} above, our DGL model captures global details like``look up at a tree house", while CLIP4Clip sees local cues, such as ``cartoon of a squid on a bike." In Fig~{\ref{visualization}} below, DGL identifies actions like ``run into the crowd" and ``catch a basketball," whereas CLIP4Clip only recognizes ``catch a basketball." Thus, the results show that DGL perceives global video information.

\begin{figure}[h!]
  \centering
  \includegraphics[width=\linewidth]{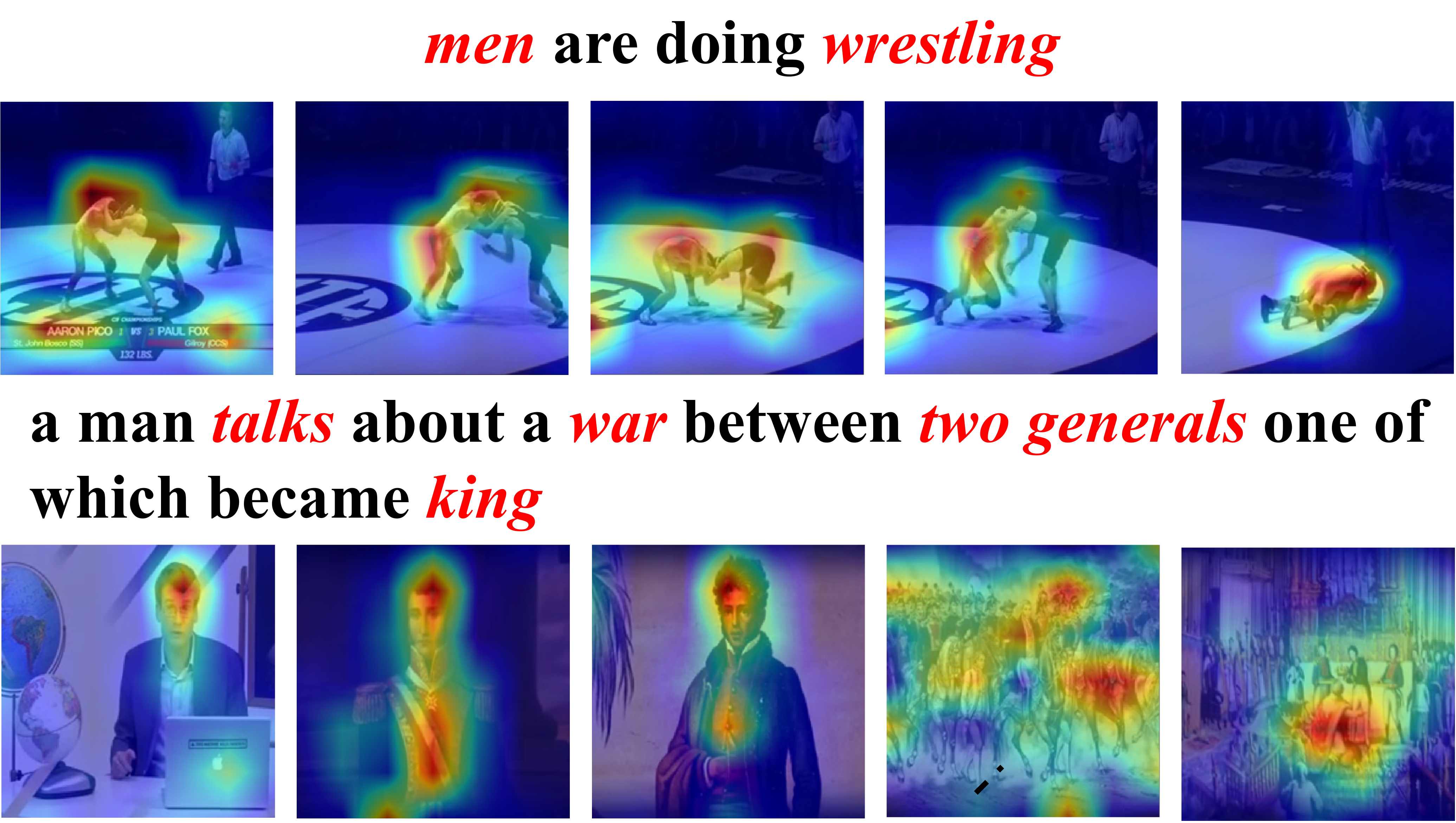}
  \caption{Visualization of global prompt, we plot the global prompt's attention weight on each frame. The red text in the query corresponds to the video's discriminative features.}
  \label{prompt}
\end{figure}

\textbf{What can global prompt learn?}
The visualized attention weights in Fig~{\ref{prompt}} reveal that the global prompt in DGL effectively captures the essence of dynamic scenes. The top figure confirms its focus on the progression of 'wrestling.' The bottom figure shows how it links discrete details—``talks,'' ``two generals,'' ``war,'' and ``king''—across frames, leading to the successful retrieval of the total video. These results demonstrate DGL's ability to capture temporal dynamics and global information from local cues.


\section{Conclusion}
In this work, we propose DGL, which generates local-level prompts for text and vision branches from a shared latent space, enhancing cross-modal interaction. 
Also, we propose a new attention mechanism for creating local and global prompts tailored to videos, which stands out in comparison to the existing literature where each frame is encoded separately by a fixed encoder.
Extensive experiments show that, compared to the fully finetuning method or naive PEFL methods, our method only trains 0.83M parameters and outperforms them on four text-video retrieval datasets. 

\section{Acknowledgments}
This work was supported in part by the Australian Research Council (ARC) under Grant DP200100938. We thank Yiyuan Yang and Shuai Zhao for their helpful discussions.


\bibliography{aaai24}

\end{document}